\pdfoutput=1
\documentclass{article}

\usepackage{microtype}
\usepackage{graphicx}
\usepackage{subfigure}
\usepackage{booktabs} 
\usepackage{amsmath,amssymb,amsfonts}

\usepackage{hyperref}

\usepackage[accepted]{icml2019}

\icmltitlerunning{Hierarchical Annotation of Images with 2AFC Metric Learning}

\begin{document}

\twocolumn[
\icmltitle{Hierarchical Annotation of Images with \\ Two-Alternative-Forced-Choice Metric Learning}



\icmlsetsymbol{equal}{*}

\author{\IEEEauthorblockN{Niels Hellinga}
\IEEEauthorblockA{\textit{Eindhoven University of Technology}\\
Eindhoven, The Netherlands \\
n.hellinga@student.tue.nl}
\and
\IEEEauthorblockN{Vlado Menkovski}
\IEEEauthorblockA{\textit{Eindhoven University of Technology}\\
Eindhoven, The Netherlands \\
v.menkovski@tue.nl}
}

\begin{icmlauthorlist}
\icmlauthor{Hellinga Niels}{equal,tue}
\icmlauthor{Menkovski Vlado}{equal,tue}
\end{icmlauthorlist}

\icmlaffiliation{tue}{Eindhoven University of Technology \\
Eindhoven, The Netherlands}

\icmlcorrespondingauthor{Menkovski Vlado}{v.menkovski@tue.nl}
\icmlcorrespondingauthor{Hellinga Niels}{n.hellinga@student.tue.nl}

\icmlkeywords{Machine Learning, ICML}

\vskip 0.3in
]



\printAffiliationsAndNotice{\icmlEqualContribution} 

\begin{abstract}

Many tasks such as retrieval and recommendations can significantly benefit from structuring the data, commonly in a hierarchical way. To achieve this through annotations of high dimensional data such as images or natural text can be significantly labor intensive. We propose an approach for uncovering the hierarchical structure of data based on efficient discriminative testing rather than annotations of individual datapoints. Using two-alternative-forced-choice (2AFC) testing and deep metric learning we achieve embedding of the data in semantic space where we are able to successfully hierarchically cluster. We actively select triplets for the 2AFC test such that the modeling process is highly efficient with respect to the number of tests presented to the annotator. We empirically demonstrate the feasibility of the method by confirming the shape bias on synthetic data and extract hierarchical structure on the Fashion-MNIST dataset to a finer granularity than the original labels.
\end{abstract}

\section{Introduction}

High-dimensional datapoints such as natural images commonly carry complex semantic information. For example to characterize an image of a clothing item it is not enough to simply label it by its type, but we also need to know its color, gender type and size. Fine-grain annotations enable many downstream task on such data. Furthermore, it allows for efficiently organizing it in a hierarchical structure \cite{Blei2004}. Therefore, a clothing e-commerce retailer may benefit from a certain hierarchical structure (i.e. gender $>$ type $>$ color $>$ size) such that its customers (or a recommendation algorithm) can find what they are looking for quicker. This is beneficial since humans naturally group and cluster similar objects together in order to form a class or super class \cite{Sternberg2008}.
As the low-level pixel information in such data is far removed from the semantic meaning and annotations we typically need complex non-linear maps build usually with deep neural networks to map to these annotations. However, train such models we also need a significant amount of annotations. 
To address this challenge we propose a method that leverages the efficiency of discrimination testing to capture the latent perception of difference between the data points by the annotators. Work in psychometics on measurement of subjective perception of objective stimuli provides strong insights in how such data collection can be effectively developed \cite{Fechner1889}. Specifically the two-alternative-forced choice (2AFC) method \cite{ehrenstein1999psychophysical}, which has been adapted for measurement of complex high-dimensional stimuli such as images and video \cite{maloney2003maximum, menkovski2012adaptive}.  
In this paper we present a method that combines 2AFC tests, with active learning methods, deep metric learning and agglomerative clustering to develop a rich embedding of the data that captures semantic relationship between the data points and uncover this semantic structure.  

In order to demonstrate the feasibility of the method we empirically confirm the shape vs color bias \cite{Ritter2017} by using our own created synthetic dataset and extract hierarchical structure on the Fashion-MNIST dataset to a finer granularity than the original labels. 

\section{Related Work}

Extracting hierarchical structure from data is a lively field of study. In \cite{Li2010}, Li et al. present two types of hierarchies studied, namely language based (i.e. WordNet \cite{Miller1995, Snow2006}) and the low level visual feature based. Even though these approaches work fine and help in tasks like image organization, they lack the visual information that connects images together. Concepts like snowy mountains and skiing are far apart from each other on the WordNet hierarchy, which is a language based hierarchical approach but visually these concepts should be closer. There have been some purely visual feature based hierarchies \cite{Ahuja2007, Bart2008} but they are difficult to interpret. There motivation comes from the fact that the authors belief that an image hierarchy is not following a language hierarchical structure. For example, sharks and whales should be close neighbours on in image hierarchy which is a useful property of tasks such as image classification. One problem of such visual hierarchies is that none of the work was able to evaluate the effectiveness directly. This is why Li et al. \cite{Li2010} created a meaningful hierarchy for end-tasks such as image annotation and classification. Given the images and their tags (labels) their approach is able to automatically create a hierarchy, which is organizes images from very general to specific attributes. Ge et al. \cite{Ge2018} propose a hierarchical triplet loss (HTL) which is able to automatically collect insightful training samples by using a predefined hierarchical structure that encodes global context information. They have two main components in their method, the constructions of the hierarchical class tree and a dynamic margin.

Fine-grained image recognition (FGIR) tasks are also closely related to extracting hierarchical structure of image data. In \cite{Lin2015} the authors introduce a bi-linear model in order to create high-order image representations which are able to compute local pairwise interactions between features of two independent sub-networks. Such approaches have been enabled by the hierarchical representation learning present in modern convolutional neural network models \cite{Chen2016, Kaiming2016}. However, due to the high dimensionality of the features it becomes impractical for subsequent analysis. In order to reduce the high dimensionality of bilinear model features, Gao et al. \cite{Gao2016} introduced a model that approximates such bilinear feature by using polynomial kernels. Kong et al. \cite{Kong2016} went a step further and introduced a classifier  co-decomposition to further restrict a bilinear model.

There has also been work that is able to capture the slight visual differences between categories \cite{Huang2016, Zhang2014} which uses bounding boxes to locate discriminative regions. The big drawback of this approach is that annotating these bounding boxes is a labour intensive process and these methods have therefore not been applicable to large-scale real world problems. In order to overcome this issue, visual attention models \cite{Chen2018, Liu2018} where applied to FGIR tasks \cite{Fu2017, Zheng2017} in order to automatically search the regions of interest. It works well since it can behave as a bounding box which where labour intensive to annotate. 
There have also been works that use extra guidance in order to learn a semantic-related regions which, in return creates a more meaningful region for FGIR tasks. Lui et al. \cite{Chen2016, Liu2017} introduced such work which makes use to part-based attribute in order to learn more discriminative features for fine-grained bird recognition. Also He et al. \cite{He2017} used detailed text descriptions in order to mine discriminative parts or characteristics.

The most recent work is that of Chen et al. \cite{Chen2018} who proposed a Hierarchical Semantic Embedding (HSE) framework which is able to predict categories of different levels in such a hierarchy and simultaneously integrate this structured correlation information which most of the other works, introduced above, overlook. Their HSE framework sequentially predicts category score vectors for each level and at each level of the hierarchy use the highest score vector as prior knowledge to learn a finer grained feature representation.

However, there are two main gaps in the above works which motivates our approach. One is that due to the labels of each data point there is a limitation to the depth of the hierarchy, meaning that non of the work shows finer granularity beyond the labels. The second is the resource intensive collection of labels in order to get a deeper hierarchy. Using our method, the embeddings allow us to extract a hierarchical structure which enables us to effectively circumvent the labour intensive process of labelling individual data points.

\section{Method}
We approach the hierarchical annotation of images by embedding the data in an embedded space that captures the semantic information that we are interested in and applying agglomerative clustering of the data in that space. 
To achieve such embedding, we use the 2AFC technique to measure the latent perception of differences by the annotators and use deep metric learning techniques to train an embedding model on these measurements. As 2AFC test can be inefficient in the number of queries to the annotator we optimize the test process by incorporating active learning techniques. 

\subsection{Two-alternative-forced-choice}
Organizing information in a hierarchical structure is a natural and efficient way for the multitude of downstream tasks that we want to enable on this data \cite{soergel1985organizing}. We expect that when presented with such data, experts or annotators use a latent structure to produce the annotations. Capturing this latent structure directly is difficult because it requires a significant effort to capture and communicate it. On the other hand, a discirminative comparisons come with much lower cost. This characteristics has been known and utilized in psychometrics specifically in measurement of subjective perception of objective stimuli \cite{Fechner1889}. More recently such methods have also been developed for measurements of subjective perception of complex stimuli such as images and video \cite{maloney2003maximum, menkovski2012adaptive, menkovski2011value}. In this work 2AFC methods have been used to efficiently capture the perception of difference in between pairs of stimuli. This allowed for modeling where an individual data point reside on a relative scale of a particular quantity. In a similar manner we use the 2AFC procedure to capture the relative difference between the pair of images for a specific question. 

\begin{figure}[]
    \centering
    \includegraphics[width=0.5\linewidth]{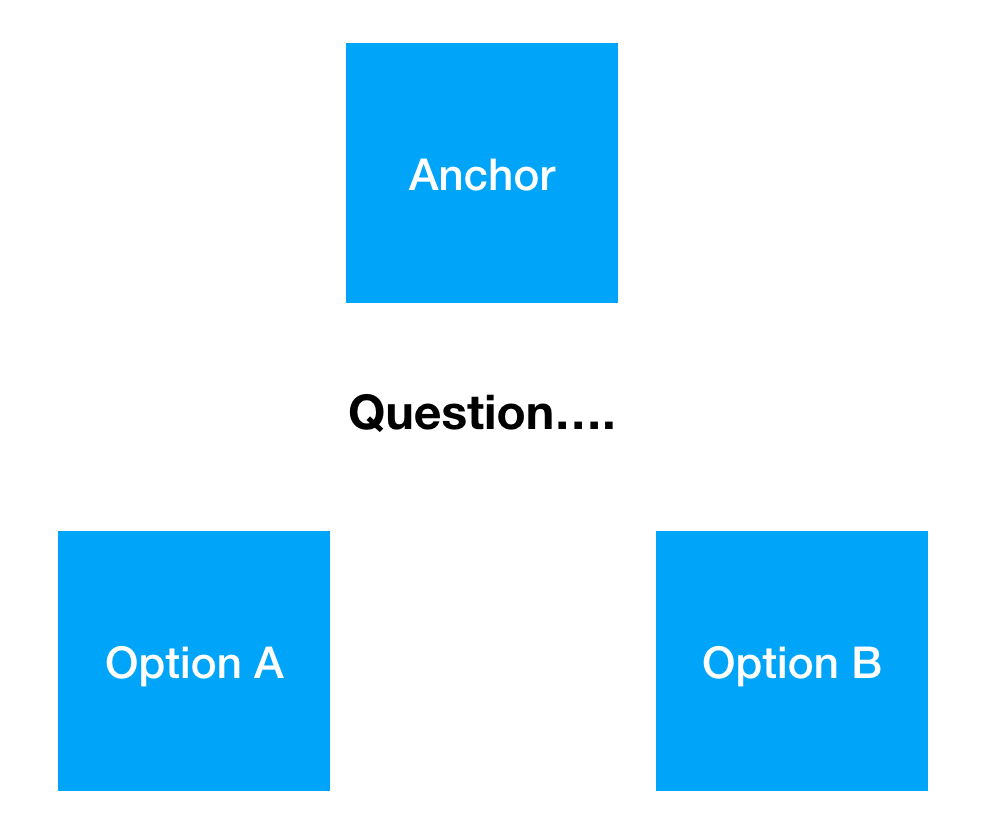}
    \caption{Triplet Selection Layout using two-alternative-forced choice method}
    \label{fig:tripletselectionlayour}
\end{figure}

As given in Figure \ref{fig:tripletselectionlayour}, we select an anchor image and two query images. We ask the annotator to discriminate between the distance given by the anchor and the first query image (option A) and the anchor and the second query image (option B). The distance is with respect to a particular quantity in the image such as: the size of the object, the category of the objects, value of the object. 
We then store the answers by marking the image which was chosen as closer to the anchor (positive) and the other as further than from the anchor (negative). 


\subsection{Deep metric learning}
Our aim is to embed the high-dimensional input data in to a space that captures the semantic structure that we want to uncover. As the input is high dimensional, we aim to rely on deep neural network models to capture the feature present in the image more effectively as demonstrated in by the advances of these methods in the image analysis domain \cite{Chen2016, Kaiming2016}. We also recognize that the input produced by the 2AFC test and our goals are perfectly aligned with the advances in deep metric learning and particularly with the triplet training procedure \cite{Schroff2015}. 

Triplet training procedure consists of three instances of the same feed forward neural network $M_e$ that share the same parameters. For this we used the highly successful ResNet model\cite{Kaiming2016}. Depending on the dataset we used a different depths of the ResNets. For images of size 128x128x3 we used a ResNet-110 and for images with size 28x28x1 we had the ResNet-20. For both experiments, the models output an embedding with a dimensionality of 8.

In order to train the model we used the loss function as given in \cite{Schroff2015}. If we define the distances with respect to the anchor ($x$) as,

\begin{align*}
 d(x,x^+) & = \vert \vert M_e(x) - M_e(x^+) \vert \vert_{2}^{2} \\
 d(x,x^-) & = \vert \vert M_e(x) - M_e(x^-) \vert \vert_{2}^{2}
\end{align*}

Then, the learning objective here is that,

\begin{align*}
    d(x,x^+) & \leq d(x,x^-) - \alpha \\
    d(x,x^+) - d(x,x^-) + \alpha & \leq 0
\end{align*}

where $\alpha$ represents the margin which enforces a distance between $d(x,x^+)$ and $d(x,x^-)$. Note that alpha is also needed such that $M_e$ cannot satisfy this equation with zero vectors for the embeddings ($M_e$(any image)). We used an alpha of $0.2$. During training the loss function will be the following:

\begin{equation}
TripletLoss = Max(  d(x,x^+) - d(x,x^-) + \alpha, 0 )
\end{equation}

\subsection{Triplet Selection}
Even though answering one of the questions is fairly quick for the annotator, the total number of available questions given a number of images is very large. Furthermore, not all questions are equally valuable for training and improving our embedding model. Such questions have been the focus of the active learning field \cite{Settles2010}. We used an active learning approach using the pool-based uncertainty sampling approach.
Algorithm \ref{alg:bayesian} shows the overall method for the active learning approach. In order to determine $Q$, we create pools of images where each pool contains close neighbours from a random selected image. From this pool of images we generate new potential questions. We can use the Bayes Factor as an uncertainty sampling method to determine if, for a given question $q_i$, whether we have a 50-50 change for choosing an answer ($a_0$ vs $a_1$) or that we have any another ratio/change such that we can be sure either $a_0$ or $a_1$ is more likely to be clicked by the annotator. Hence, we would like to compare two similar models for $a_0 \sim Bin(n,\Theta)$ given that model $M_1$ has a $\Theta = 0.5$ and model $M_2$ has an unknown $\Theta$. For $M_2$ we will take the prior distribution for $\Theta$ to be uniform on $[0,1]$.

Using Bayes Factor we can construct the following likelihood ratio $BF = \frac{P(N'_i|M_1)}{P(N'_i|M_2)}$ where $N'_i$ is the set of all neighbouring questions to $q_i$. If $BF > 1$ then we can strongly assume, given the data $N'_i$, that $M_1$ is supported over $M_2$. Any value of $BF < 1$ we can assume that $M_2$ is supported by the data. In our case, if $BF < 1$ then we can assume that we know either $a_0$ or $a_1$ will be clicked by the annotator and that we do not need to ask this question again.

In order to calculate $BF$ we need to know $P(N'_i|M_1)$ and $P(N'_i|M_2)$.

\begin{align*}
    P(N'_i|M_1) &= {n \choose k} \Theta^k (1-\Theta)^{n-k} \\
    &= {n \choose k} 0.5^k (1-0.5)^{n-k} \\
    &= {n \choose k} 0.5^n \\
    P(N'_i|M_2) &= \int_{0}^{1}{n \choose k}  \Theta^k (1-\Theta)^{n-k} d\Theta \\
    &= {n \choose k} \int_{0}^{1} \Theta^k (1-\Theta)^{n-k} d\Theta \\
    &= {n \choose k} B(k+1, n-k+1) \\
    &= {n \choose k} \frac{\Gamma(k+1)\Gamma(n-k+1)}{\Gamma(k+n-k+2)} \\
    &= \frac{n!}{k!(n-k)!} \frac{k!(n-k)!}{(k+n-k+1)!} \\
    &= \frac{n!}{(n+1)!} \\
    &= \frac{1}{n+1}
\end{align*}

where $n$ is the total amount of clicks and $k$ is equal the amount of $a_0$ clicks. Note that it does not matter if we count $a_0$ or $a_1$ since the test here is whether the model 'guesses' or not. If either of the two answers is favoured then $M_2$ will be supported by $N'_i$. Knowing $P(N'_i|M_1)$ and $P(N'_i|M_2)$ we can calculate BF as,

\begin{align*}
    BF &= \frac{P(N'_i|M_1)}{P(N'_i|M_2)} \\
    &= \frac{{n \choose k} 0.5^n}{1/(n+1)} \\
    &= {n \choose k} 0.5^n (n+1)
\end{align*}

The main idea is that we want to know if, given a current question and all the previous answers, whether we probability of clicking an answer will be a 50\% change or not. If there is a high probability of choosing any of the two answers we do not need to ask the question. Whereas, if the probability of choosing an answer is 50\% then we need to ask the question to the annotator since we cannot be sure yet. 

\begin{algorithm}[tb]
  \caption{Triplet selection}
  \label{alg:bayesian}
\begin{algorithmic}
  \STATE Initialize 
  \STATE $T = \{q_1, q_2,...,q_{n}\}$ - set of $n$ random unanswered triplets.
  \STATE $D = \{\}$ - set of answered triplets
  \STATE $\tau = 0.75$
  \WHILE{not converged}
  \STATE $D \gets$ Have annotators answer $T$
  \STATE Update $M_e$ with $D$
  \STATE $Q \gets $ select new potential questions
  \STATE $T = \{\}$
  \FOR{$q$ in $Q$}
  \IF{$BF(q) > \tau$}
  \STATE $T \gets$ add $q$
  \ENDIF
  \ENDFOR
  \STATE Sort $T$ by highest $BF$
  \STATE $T \gets$ top 0.8 triplets of $T$ + 0.2 random triplets for generality
  \ENDWHILE
\end{algorithmic}
\end{algorithm}

\subsection{Agglomerative clustering}
After the utility of asking further questions to the annotators has diminished we conclude that we can now successfully embed the data such that its semantic information is captured by the distance metric of the space. To extract this information we run a complete-linkage agglomerative clustering algorithm \cite{rokach2005clustering} and produce a dendrogram that represents the captured structure. 

\section{Experiments and results}

To evaluate the proposed method we develop two empirical studies. In the first one we test whether the method can uncover the well studied shape bias in humans \cite{landau1988importance} on a synthetic dataset. In the second we extract hierarchical structure on the FashionMNIST dataset \cite{Xiao2017} containing images of clothing items. 

\subsection{Shape bias on simple shapes}
\begin{figure}[]
    \begin{minipage}[b]{.5\linewidth}
        \centering
        \includegraphics[width=0.5\textwidth]{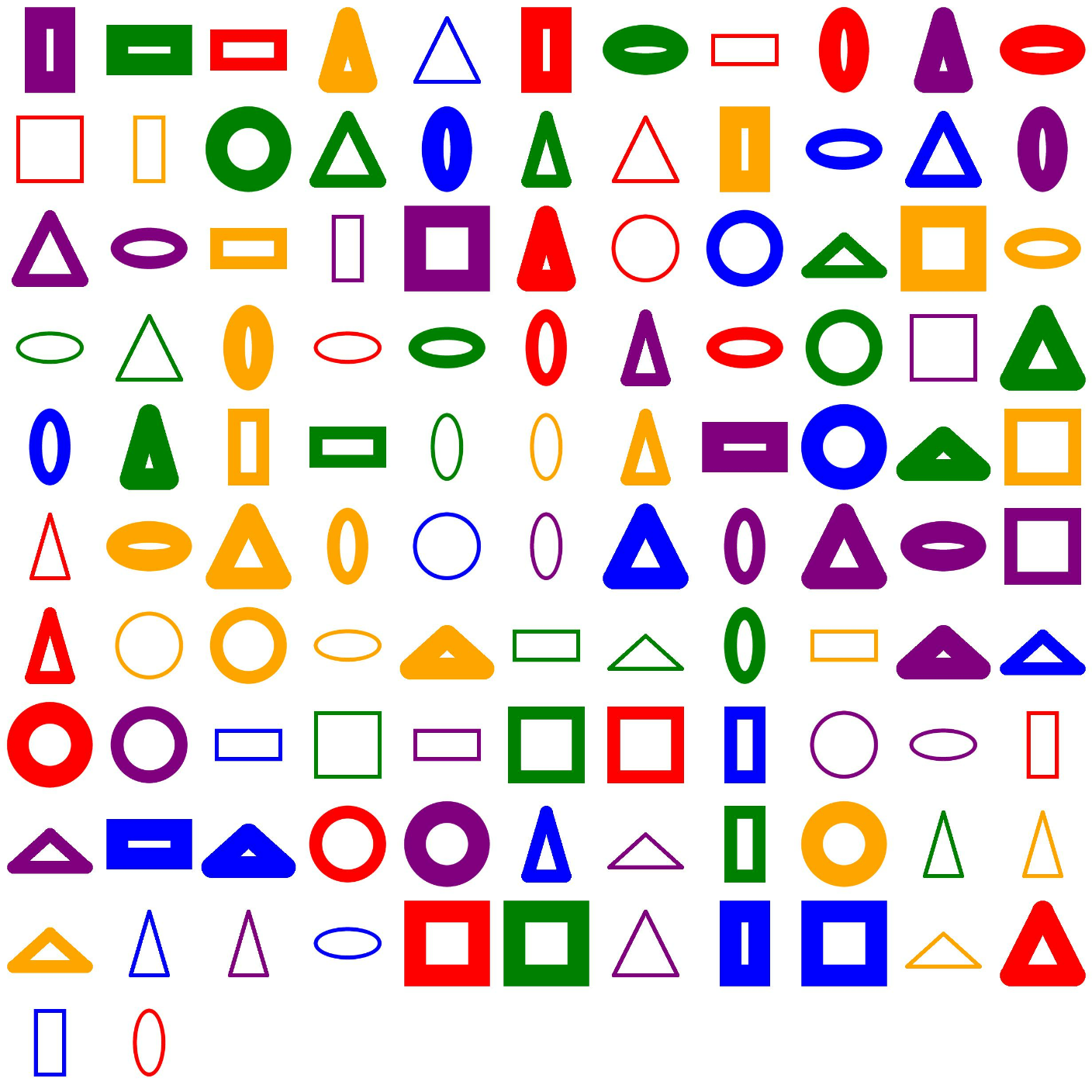}
    \end{minipage}
    \hspace{1cm}
    \begin{minipage}[b]{.2\linewidth}
        \centering
        \includegraphics[width=0.7\textwidth]{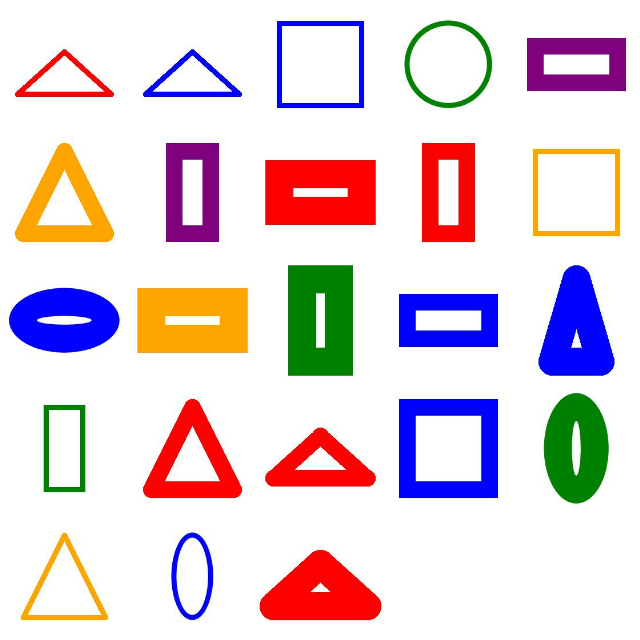}
    \end{minipage}
    \caption{Simple Shape - train set left and test set right}
    \label{fig:simpleshapes}
\end{figure}

We have created a synthetic simple-shapes dataset which contains 9 different shapes where each shape has 3 different thicknesses and each shape and thickness has 5 different colors. Hence, 135 unique objects that we split into a train and test set (Figure \ref{fig:simpleshapes}). The dimensionality of the images is 128x128x3. In this experiment the annotators give answer to the question "\textit{Which object is more similar to the anchor object?}". 

\begin{figure}[]
    \centering
    \includegraphics[width=1\linewidth]{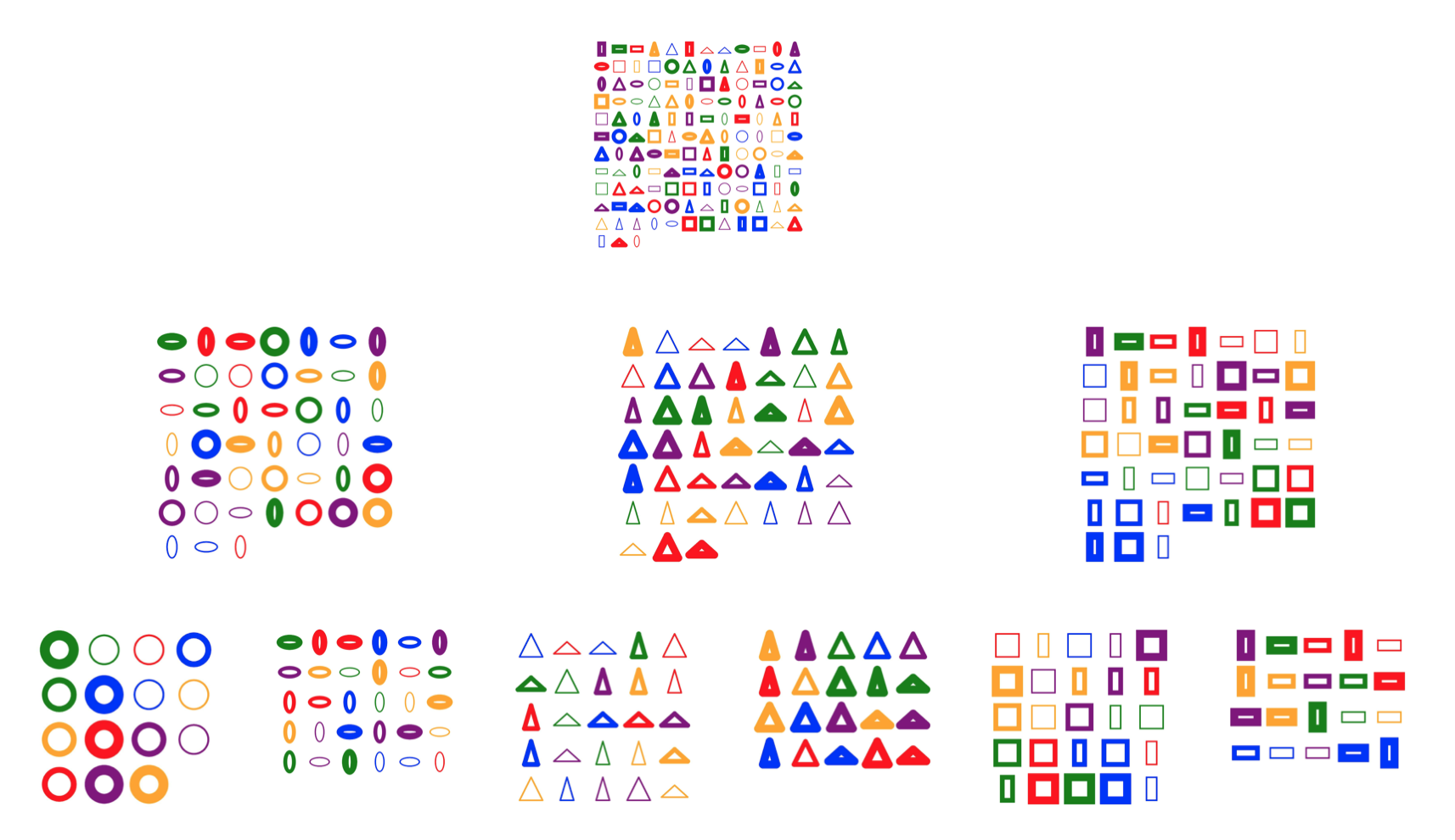}
    \caption{Simple Shape dendrogram splits}
    \label{fig:2afc_ss_dendogram}
\end{figure}

After collecting 840 triplets, we trained the ResNet-110 model with the specified triplet loss and extracted the data structure using the complete-linkage clustering \cite{defays1977efficient} algorithm. 
Figure \ref{fig:2afc_ss_dendogram} shows the resulting splits. We can clearly see that the resulting clusters are based on the shape and not color or thickness of the objects in the images. The initial three spits are  separating the different shapes: circles, triangles and rectangles. The next level the shape is again is the discriminator for the case of the circles and the rectangles, while in the case of the triangles the results are not as clear. This is somewhat expected as the case of the triangle height of the triangle is not connected to a different concept as in the case of the circle vs. oval. In the case of the rectangles the squares and the vertical rectangles are clustered against the horizontal rectangles. 


%
%


\subsection{Fashion-MNIST}
Using the 2AFC metric learning method, we are also able to extract a hierarchical structure based on the perception of difference of the annotator. We will be using the Fashion-MNIST dataset \cite{Xiao2017} with the question "\textit{Which object looks more similar to the anchor object?}".

\begin{figure}[]
\centering
    \begin{minipage}[b]{1\linewidth}
        \centering
        \includegraphics[width=1\linewidth]{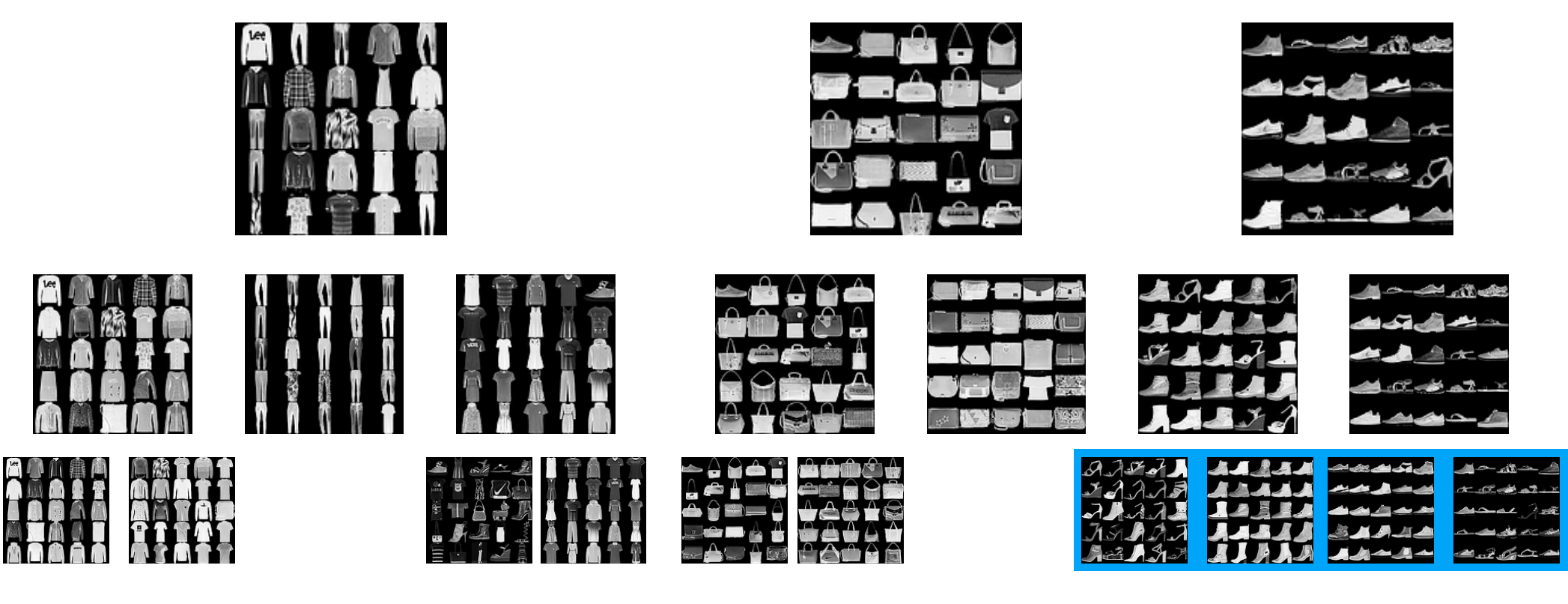}
        \caption{Fashion-MNIST initial splits}
        \label{fig:fashionmnis_initialSplit}
    \end{minipage}
    \begin{minipage}[b]{1\linewidth}
        \centering
        \includegraphics[width=1\linewidth]{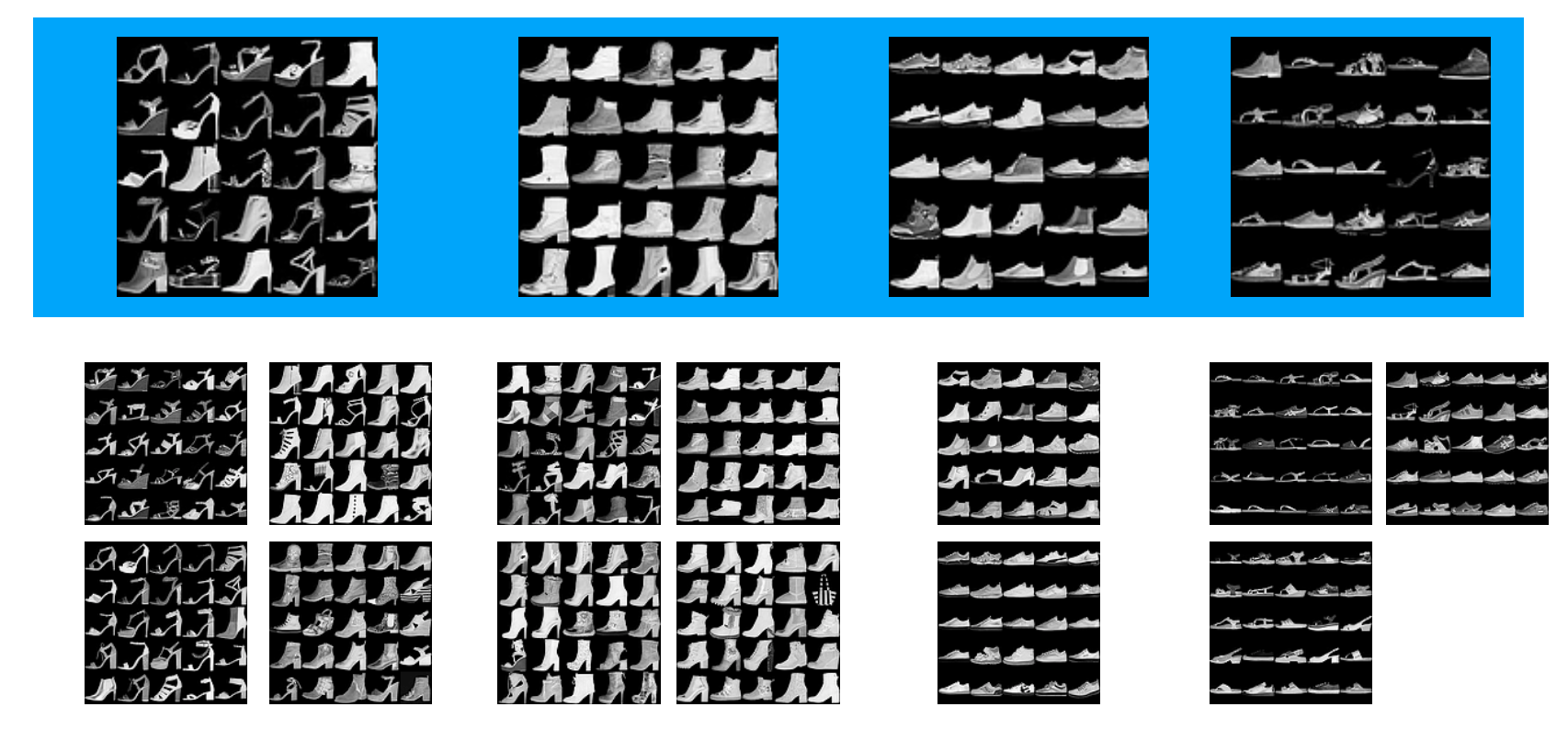}
        \caption{Fashion-MNIST granularity - blue}
        \label{fig:fashionmnis_blue}
    \end{minipage}
\end{figure}

%
%
Results of the initial splits can be seen in Figure \ref{fig:fashionmnis_initialSplit}. Note that we can clearly see that the first split is based on cloth (left), bags (middle) and shoes (right) which continues further down in more fine-grained detail. Further splits of shoes can be seen in Figure \ref{fig:fashionmnis_blue}. Here we can clearly see that we end up with clusters that present us with a finer granularity than the original Fashion-MNIST labels. We can observe for example that sandals have been split into high-heal sandals and flat sandals. In order to construct this hierarchical structure we used 1700 triplets.

We further contrast these results with clustering on the raw pixel values to form a baseline and demonstrate the value of developing the embedding space using the 2AFC tests. To compare the two sets of clusters we compute the normalized mutual information. Both results are then compared to the true labels of the Fashion-MNIST dataset. Results can be found in Table \ref{tab:nmi_Fashion-MNIST}. Note that 'Level' is based on a binary tree level and therefore the nodes are the amount of clusters created at each level. 

The results demonstrate empirically that the 2AFC method produces an embedding in which clustering captures the semantic structure in the data. We also show that our proposed method allows us create clusters with finer granularity than the dataset labels.

\begin{table}[]
\centering
\caption{Normalized Mutual Information compared to true labels given}
\label{tab:nmi_Fashion-MNIST}
\begin{tabular}{|l|l|l|}
\hline
Level & Baseline & 2AFC  \\
\hline
0     & 0.000    & 0.000 \\
1     & 0.192    & 0.392 \\
2     & 0.345    & 0.491 \\
3     & 0.426    & \textbf{0.583} \\
4     & 0.487    & 0.562 \\
5     & \textbf{0.499}    & 0.520 \\
\hline
\end{tabular}
\end{table}

\section{Conclusion}
In this work we present a method that leverages the efficiency of discrimination 2AFC testing using to capture the latent perception of difference between data points. We have shown that we are able to capture the shape bias with synthetic data and have shown that it is possible to extract a meaningful hierarchical structure on the Fashion-MNIST dataset, resulting in a finer granularity than the original labels. We have also achieved this efficiently by incorporating an active learning triplet selection based on Bayesian Factor estimation. 

There are wide variety of applications that can benefit from extraction of hierarchical structure of data both in imaging domains such as medical imaging, but also broader in other domains that rely on high dimensional datasets. 


\bibliography{ref}
\bibliographystyle{icml2019}

\end{document}